%% file: main.tex
\documentclass[10pt,twocolumn,letterpaper]{article}

\usepackage{cvpr}              % To produce the CAMERA-READY version
\input{preamble}

\definecolor{cvprblue}{rgb}{0.21,0.49,0.74}
\usepackage[pagebackref,breaklinks,colorlinks,citecolor=cvprblue]{hyperref}
\usepackage[accsupp]{axessibility}  % Improves PDF readability for those with disabilities.

\def\paperID{11} % *** Enter the Paper ID here
\def\confName{CVPR}
\def\confYear{2024}

\title{Learning Surface Terrain Classifications from Ground Penetrating Radar}

\author{Anja Sheppard, Jason Brown, Nilton Renno, and Katherine A. Skinner \\
The University of Michigan \\
2505 Hayward St, Ann Arbor, MI 48109\\
{\tt\small anjashep@umich.edu, jaybrow@umich.edu, nrenno@umich.edu, kskin@umich.edu}
}

\begin{document}
\maketitle
\input{sec/0_abstract}    
\input{sec/1_intro}
\input{sec/2_background}
\input{sec/3_technicalapproach}
\input{sec/4_expandresults}
\input{sec/5_discussion}
\input{sec/6_conclusion}
\input{sec/7_acknowledgements}
{
    \small
    \bibliographystyle{ieeenat_fullname}
    \bibliography{main}
}

% WARNING: do not forget to delete the supplementary pages from your submission 
% \input{sec/X_suppl}

\end{document}

%% file: preamble.tex
\pdfoutput=1
\usepackage[dvipsnames]{xcolor}

\usepackage{tabularray}
\usepackage{pgfplots}
\usepackage{float}

\pgfplotsset{compat=1.18}

\usepackage{makecell}

%% file: sec/0_abstract.tex
\begin{abstract}
Terrain classification is an important problem for mobile robots operating in extreme environments as it can aid downstream tasks such as autonomous navigation and planning. While RGB cameras are widely used for terrain identification, vision-based methods can suffer due to poor lighting conditions and occlusions. In this paper, we propose the novel use of Ground Penetrating Radar (GPR) for terrain characterization for mobile robot platforms. Our approach leverages machine learning for surface terrain classification from GPR data. We collect a new dataset consisting of four different terrain types, and present qualitative and quantitative results. Our results demonstrate that classification networks can learn terrain categories from GPR signals. Additionally, we integrate our GPR-based classification approach into a multimodal semantic mapping framework to demonstrate a practical use case of GPR for surface terrain classification on mobile robots. Overall, this work extends the usability of GPR sensors deployed on robots to enable terrain classification in addition to GPR's existing scientific use cases.
\end{abstract}

%% file: sec/1_intro.tex
\section{Introduction}
\label{sec:intro}

\begin{figure}
    \centering
    \includegraphics[width=\linewidth]{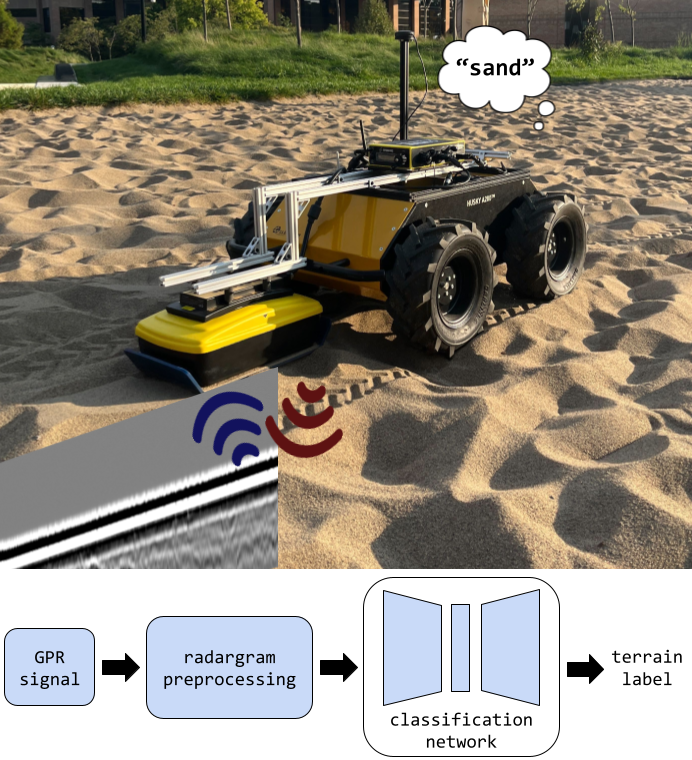}
    \caption{Clearpath Husky robot with a GPR mounted on the back. The radargram, which represents the permittivity of the surface and subsurface materials sensed by the GPR as a time series, is shown in image form in the bottom left.}
    \label{fig:radar}
\end{figure}

The Ground Penetrating Radar (GPR) is an instrument with a long history in non-intrusive surveying of sub-surface features with applications such as transportation infrastructure mapping \cite{Lalague2015}, cemetery grave localization \cite{Fielder2009}, and geologic feature identification \cite{Eder2008}. The reflected signal returned from the radar pulse, shown in \cref{fig:radar}, encodes information about the velocity of electromagnetic waves through materials below the sensor. Recent literature has shown use cases for GPR to characterize surface properties, such as soil water content \cite{Wu2019}. Additionally, GPR has been gaining interest in the robotics community for light-invariant autonomous vehicle localization \cite{Cornick2016}.

Terrain classification is a challenging problem for mobile robots in extreme environments. One challenge is that training RGB segmentation networks generally requires a lot of hand-labeled, diverse data. Additionally, camera-based approaches can suffer when terrain is occluded or the lighting is poor \cite{Milford2014}. This has led to an interest in alternative sensor modalities for classifying terrain, such as LiDAR \cite{Shaban2021}, Inertial Measurement Units (IMUs) \cite{Guan2022}, torque sensors \cite{Brooks2012}, and microphones \cite{Zurn2019}.

As GPR sensors are already being deployed on mobile robots in remote environments to accomplish science tasks -- such as the RIMFAX \cite{Hamran2020} radar on the Perseverance rover -- we seek to determine whether this sensor can be leveraged to aid terrain classification. This paper seeks to answer the following question: Can GPR distinguish \textit{surface} terrains? In our quest to answer this, we present the following contributions:
\begin{enumerate}[label=(\roman*)]
    \item We investigate the use of GPR direct waves for learning-based surface terrain classification. 
    \item We conduct field experiments with a custom robot platform to collect a real dataset containing overlapping RGBD images, GPR radargrams, and GPS coordinates.
    \item We develop a framework that takes input GPR data and outputs terrain classifications, and we present qualitative and quantitative results to evaluate this framework.
    \item We integrate GPR terrain classification into a multimodal framework for terrain mapping to motivate future work.
\end{enumerate}

The remainder of this paper is organized as follows: we first discuss standard uses for the GPR and prior work with other sensing modalities for terrain classification in \cref{sec:background}. In \cref{sec:technicalapproach}, we present our formulation of GPR-based classification. In \cref{sec:expandresults}, we share details about our robot data collection platform and we demonstrate experimental results on our GPR terrain classification networks. Additionally, we present preliminary results of a multimodal terrain mapping framework that fuses GPR and RGB data. Finally, in \cref{sec:discussion}, we discuss future work motivated by this novel application of GPR.

%% file: sec/2_background.tex
\section{Background}
\label{sec:background}

\subsection{Terrain Classification}

Terrain classification is an important and well-studied problem for autonomous robots operating in extreme environments, such as the Mars rovers \cite{Ono2015}. Prior work on terrain classification has focused primarily on semantic segmentation for RGB and depth images. However, some recent works also demonstrate novel uses of unconventional sensors in multimodal frameworks for learning terrain.

\subsubsection{RGB Camera and Depth Sensor Approaches}

Exteroceptive sensors, such as cameras and range sensors, provide context about the robot's environment from a distance. The Soil Property and Object Classification (SPOC) network proposed by NASA \cite{Atha2022} presents a lightweight, vision-based approach to the terrain classification problem for Mars rovers. Another approach aims to provide more fine-grained labels of RGB rover images that are useful for science-based navigation goals by focusing particularly on terrain texture \cite{Panambur2022}. In order to extend advances in Martian terrain classification, \cite{Vincent2022} uses unsupervised contrastive pretraining to improve generalization to sites on Mars that were not present in the image training set.

LiDAR, while typically a sensor used to provide geometric information about a scene, has also proven useful for classifying semantics, as demonstrated in \cite{Shaban2021}. However, LiDARs are resource-intensive, making them difficult to include on planetary rovers or other resource-constrained mobile robots.

Vision-only approaches to terrain classification suffer from several sensor-specific weaknesses: lighting, weather, and the potential for obstacles to be obscured can cause unreliable results in the real world. Additionally, vision-based segmentation models often require tedious pixel-wise labeling. The AI4MARS dataset released by NASA's Jet Propulsion Laboratory \cite{Swan2021} relied on crowdsourcing volunteers to label over 35,000 images. Multimodal approaches can help compensate for weaknesses in unimodal RGB-based terrain classification and can also potentially reduce the burden of manual image labeling.

\subsubsection{Multimodal Approaches}

Some works have leveraged additional sensor modalities to supervise the training of vision-based terrain classification networks. One foundational study uses a traversability metric calculated from wheel torque sensors to supervise the labeling of images \cite{Brooks2012}. Similarly, \cite{Zurn2019} presents a multimodal model that learns a deep clustering encoder for audio data of the wheel/terrain interaction and then produces pixel-wise semantic labels for the corresponding forward-facing camera images.

Recent approaches have also focused on sensor fusion within a multimodal network architecture, such as \cite{Guan2022}, which fuses visual and inertial data via a shared embedding space to detect both known and unknown terrain types. Additionally, \cite{Ishikawa2021} proposes the use of Multimodal Variational Autoencoders (VAEs) and Gaussian Mixture Models to extract features from audio and visual data into a shared embedding space. This approach allows for a more robust framework that still performs well when one of the modalities is missing.

In this work, we aim to add GPR to the growing list of sensors that can classify terrain. We also hope to demonstrate its potential in compensating for weaknesses of unimodal RGB approaches through a multimodal terrain mapping framework.

\subsection{Ground Penetrating Radar}
Over the past few decades, autonomous robots with GPR have been deployed in remote environments for various tasks including subsurface infrastructure reconstruction \cite{Feng2023}, crevasse mapping for safe traversal \cite{Williams2012}, and water detection on Mars \cite{Hamran2020, Herve2020}.

GPR is typically used to investigate subsurface features \cite{Peters1994}. In recent years, some works have explored radar-based property mapping, such as GPR-mounted drones for mapping soil moisture \cite{Wu2019}, identifying the extent of subway deterioration \cite{Dawood2020}, and reconstructing underground tree roots beneath streets \cite{Lantini2020}. Due to the generally static nature of subsurface features, GPR has also demonstrated usefulness in aiding loop closure for autonomous vehicle localization \cite{Cornick2016}. Datasets recently released by Carnegie Mellon University \cite{Baikovitz2021} and the Massachusetts Institute of Technology \cite{Ort2021} are explicitly aimed at accelerating research in the use of GPR for spatio-temporal mapping.

To the best of our knowledge, no one has extended the usage of GPR to characterizing surface terrain properties for the purpose of terrain classification. A unique property of GPR called the direct wave (discussed in \cref{sec:gpr}) allows this sensor to be used for surface characterization in addition to scientific objectives. The nature of the GPR sensor model makes classification based on this metric light-agnostic, allowing it to be a suitable complement for vision-based terrain identification. In this paper, we begin to unlock a new dimension of GPR use cases for robotic navigation in extreme environments by demonstrating surface terrain classification with GPR data.

%% file: sec/3_technicalapproach.tex
\section{Technical Approach}
\label{sec:technicalapproach}

Within this section, we detail the key technical components of our approach to GPR terrain classification. Figure \ref{fig:radar} shows an overview of our proposed framework. Our framework takes GPR data as input and performs radargram preprocessing. The preprocessed data is passed into a neural network to output a terrain label. Each component of this framework is detailed further below.

\begin{figure}[htbp]
\centering
    \begin{subfigure}{.45\linewidth}
      \centering
      \includegraphics[width=\linewidth]{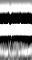}
      \caption{Asphalt}
    \end{subfigure}
    \hfill
    \begin{subfigure}{.45\linewidth}
      \centering
      \includegraphics[width=\linewidth]{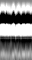}
      \caption{Grass}
    \end{subfigure}
    \begin{subfigure}{.45\linewidth}
      \centering
      \includegraphics[width=\linewidth]{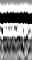}
      \caption{Sand}
    \end{subfigure}
    \hfill
    \begin{subfigure}{.45\linewidth}
      \centering
      \includegraphics[width=\linewidth]{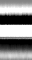}
      \caption{Sidewalk}
    \end{subfigure}
\caption{60 x 32 pixel slice of a GPR radargram direct wave for four terrain types: a. asphalt, b. grass, c. sand, and d. sidewalk.}
\label{fig:radargram}
\end{figure}

\subsection{GPR Signal Characterization}\label{sec:gpr}

GPR uses radio waves to probe subsurface composition in low loss dielectric material areas. The complex permittivity, $\varepsilon$, expresses how well a material responds to electric flux \cite{Cole1941}:
\begin{equation}
    \varepsilon = \varepsilon' - i \cdot \varepsilon''
\end{equation}
The real component, $\varepsilon'$, quantifies the stored energy, and the imaginary component, $\varepsilon''$, quantifies the dielectric loss of the material.

The relative permittivity constant, $\kappa$, also called the dielectric constant, is often expressed as the ratio between the material's permittivity and that of a vacuum ($\varepsilon_0)$ \cite{Jol2009}:
\begin{equation}
    \kappa = \frac{\varepsilon}{\varepsilon_0}
\end{equation}
Materials such as wet sand, clay, or pockets of groundwater have high relative permittivity, making signal penetration much more challenging. This is reflected in a high value of the imaginary portion of the dielectric constant, strongly damping the GPR signal.

The amount of the radio signal that is reflected at a boundary between substances is a ratio of the magnitude of the dielectric constants of both materials, $\kappa_1$ and $\kappa_2$, and it is called the reflection coefficient ($R$) \cite{Knight2001}:
\begin{equation}
    R = \frac{\sqrt{\kappa_1} - \sqrt{\kappa_2}}{\sqrt{\kappa_1} + \sqrt{\kappa_2}}
\end{equation}
The higher the reflection coefficient, the stronger the reflection. For example, when the radar signal passes from a high permittivity material, such as water, to a low permittivity material, such as dry sand, there is a positive reflection coefficient. Higher reflectivity at a boundary results in a more visible marker in the radargram (see the contrasting light and dark returns in \cref{fig:radargram}). Each vertical line of pixels in the radargram corresponds to one radar ping return from the GPR, where the returns are in mV.

The depth $d$ of the features shown in a radargram can be calculated by:
\begin{equation}
    d = v \times \frac{t}{2}
\end{equation}
where $v$ is the wave travel velocity of the ground material (a value dependent on the soil makeup) and $t$ is the two-way travel time \cite{Xie2021}. Calculating the depth is very difficult unless wave velocity is well-characterized or if the depth to a feature is manually measured. However, this tells us that the pixels towards the top of the radargram represent radiowave returns that were received the fastest, and thus represent terrain that was closer to the sensor.

\begin{figure}
    \centering
    \includegraphics[width=\linewidth]{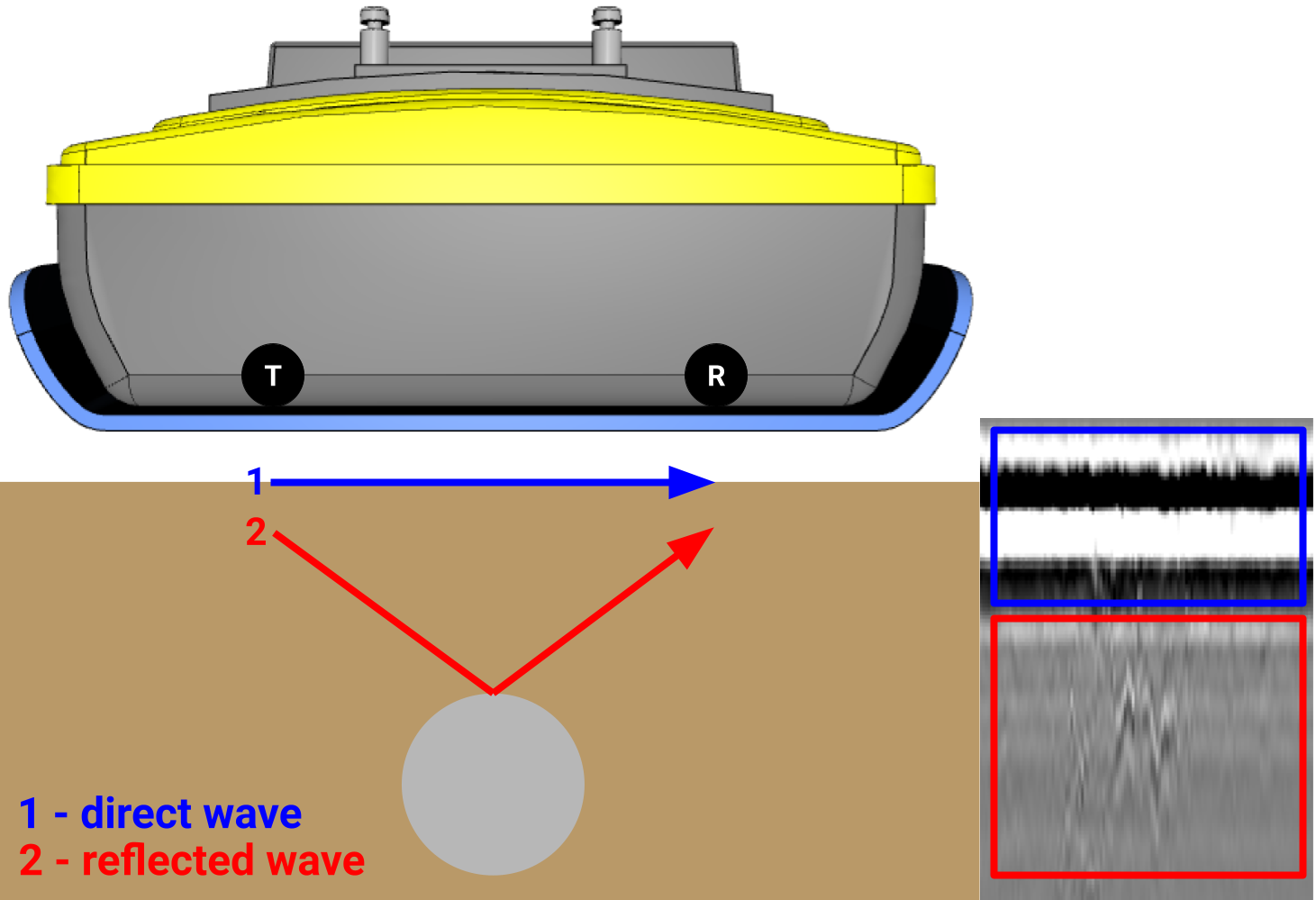}
    \caption{The radar propagation from GPR pulses has a unique property where it travels through the air and the surface of the ground (direct wave, blue), and also through the subsurface material itself (reflected wave, red). The radargram on the right shows how these different received waves appear in a GPR radargram.\vspace{-0.53em}}
    \label{fig:direct-wave}
\end{figure}

An important property of GPR is the direct wave, which is an effect of the radar signals travelling directly through the air and top surface of the ground to the receiver. See \cref{fig:direct-wave} for a visual comparison of direct and reflected waves. In many cases where GPR is being used to characterize subsurface features such as pipes or geologic formations, the direct waves are removed with background subtraction filters \cite{Sharma2017}. However, in this work we are especially focused on the surface terrain, making the direct waves of particular interest. Figure \ref{fig:radargram} shows us the qualitative visual distinctiveness of the direct waves for the four terrains we have focused on in this work. In \cref{sec:exp}, we directly investigate how the network performs on both the direct wave and the reflected wave in order to support our use of the direct wave phenomena for surface terrain classification.

The frequency of the GPR dictates its penetration depth and resolution \cite{Jol2009}. A low frequency GPR can penetrate hundreds of meters underground, but with a lower resolution. As we are interested primarily in surface and shallow subsurface terrain information, we chose a high frequency 500 MHz GPR sensor in order to benefit from higher resolution data.

\subsection{Radargram Preprocessing}

As discussed in \cref{sec:gpr}, a GPR sensor returns a vector for each sample that represents the electromagnetic wave reflectivity of materials below the sensor. These returns are then concatenated into an image representation called a radargram. In order to create our training image set, we first pad each radargram from $h \times w$ into $h \times w_{\text{pad}}$ according to the following equation, where $\%$ represents the modulus operator:
\begin{equation}
    w_{\text{pad}} = w + ((w - w_{\text{resize}}) \% s)
\end{equation}
Next, we slice horizontally with resize width, $w_{\text{resize}}$, along the area of the image representing the direct wave at an interval of $s$, the stride length. For our experiments, we used a stride length of 4 pixels.

As each radargram slice represents multiple samples from the GPR, there is the chance for more than one terrain to be represented in a sliced image. This will certainly happen in the real world as the sensor traverses terrain boundaries. For the purposes of these experiments, the radargrams were sliced to isolate each terrain into separate images before further processing into the test and train sets.

The radargrams were split with an 80/20 train/test split before being further sliced into widths 1, 8, 16, 24, and 32 pixels for use in our experiments. This was done in order to provide a frozen test set across all experiments.

\subsection{Classification Networks} \label{sec:classification}

We investigated four neural network architectures for their effectiveness in classifying terrain from GPR data. Each of the networks takes in GPR signals -- either concatenated into a 2D image or just each individual 1D sample vector. The output of each network is a single class prediction: asphalt, grass, sand, or sidewalk. Once a terrain class for a segment of GPR samples is predicted, this can be integrated into a terrain mapping framework as shown in \cref{sec:mapping}.

\subsubsection{AlexNet}
The AlexNet \cite{Alex2012}, first proposed in 2012, is a common supervised approach to RGB image classification. Although GPR radargrams are not images in the RGB space, each vector return from the sensor can be concatenated into a 2D radargram image for training and inference. The input to the AlexNet is this 2D grayscale radargram image, as shown in \cref{fig:radargram}.

\subsubsection{ResNet101}
ResNet \cite{He2015} is typically viewed as an improvement upon AlexNet in the supervised image classification problem domain. We include ResNet101 as a supervised baseline for radargram classification. Similarly to the AlexNet, the input to the ResNet network is a 2D grayscale radargram image.

\subsubsection{1D Convolutional Neural Network}
One consequence of formulating GPR data as a 2D image is that a real-time terrain classification system must wait for multiple data points to be obtained before being deployed. Additionally, it is challenging to determine the boundary between two terrain types present in one radargram slice, as the labels are per trace sample. This can be alleviated by evaluating signal-based networks that take in a single 1D GPR trace for classification. A common approach to 1D signal classification is the 1D Convolutional Neural Network (CNN) \cite{Kiranyaz2021}, which takes as an input the (200 $\times$ 1) GPR sample return vector and outputs a class prediction.

\subsubsection{Variational Autoencoders for Deep Clustering}
VAEs are unsupervised networks designed to encode an image into a 1D latent space, and then reconstruct that image from the encoding \cite{Kingma2013}. VAEs can be useful for learning underlying embedded structures in an image \cite{Song2013}. In our particular approach, modeled after \cite{Guo2017}, we modify the standard VAE reconstruction loss function to include incentive for better clusters. After performing k-means clustering on the embedding space, we can use the autoencoder to predict the terrain class of the test set.

%% file: sec/4_expandresults.tex
\section{Experiments and Results}
\label{sec:expandresults}

\subsection{Mobile Robot Platform}

\begin{figure}[t]
    \centering
    \includegraphics[width=\linewidth]{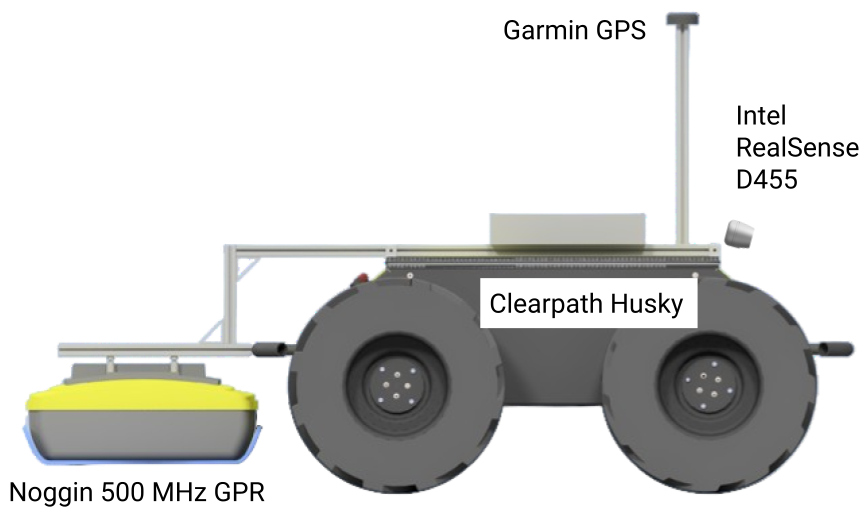}
    \caption{Our custom Clearpath Husky robotic sensor platform equipped with a GPR, a GPS, an RGBD camera, and wheel encoders.}
    \label{fig:husky_sensor}
\end{figure}

A mobile robot platform (shown in Fig. \ref{fig:husky_sensor}) was used for collecting data and deploying the GPR classification networks explored in this work. We used a Clearpath Husky robot running off of a Jetson Nano with Ubuntu 19.04 and ROS Melodic. The Husky robot has built-in wheel encoders for use in filtered odometry data calculation. Our modifications to the platform included custom mounts for a Noggin 500 MHz GPR, a Garmin GPS module, and an Intel Realsense D455 RGBD camera. The 500 MHz GPR can penetrate 5 - 10 meters below the surface (depending on the material) with a high resolution, making it the most suitable for the task of surface terrain classification compared to lower frequency sensors. A custom ROS driver for the GPR was also developed as a part of this work.

\subsection{Data Collection}
Data collection occurred over several months from July through November 2023, in temperatures ranging from 40$^\circ$ F to 70$^\circ$ F. Table \ref{tab:field_sites} details the specific locations in southeast Michigan and at the University of Michigan where different terrain data was collected. At each site, the radargram was processed for training by slicing along terrain boundaries, if any were present. This was done by cross-referencing camera images of the site. A total of 64 radargram images was collected of varying lengths. This amounts to 7752 individual traces across all radargrams and terrain categories. Some data collection was done following days of precipitation. This is important to note as radio wave propagation is heavily impacted by water.

\begin{table}[H]
    \centering
    \begin{tabular}{|l|l|}
        \hline
        Field Site & Terrain Type Collected \\
        \hline
        Independence Lake, MI & asphalt, grass, sand \\
        UM North Campus & asphalt, grass, sand, sidewalk \\
        UM MCity & asphalt \\
        \hline
    \end{tabular}
    \caption{Field data collection sites}
    \label{tab:field_sites}
\end{table}

\subsection{Terrain Classification Networks}

As described in \cref{sec:classification}, we investigated four different networks for classifying GPR radargrams: AlexNet \cite{Alex2012}, ResNet101 \cite{He2015}, a 1D CNN \cite{Kiranyaz2021}, and an autoencoder for deep clustering \cite{Guo2017}. All networks were trained on an NVIDIA GeForce RTX 3090 with 24 GB of memory. AlexNet and ResNet101 were pretrained on ImageNet, trained until convergence at 20 epochs, and employed cross entropy loss. The 1D CNN was also trained until convergence at 20 epochs. We stopped training after 20 epochs to avoid overfitting. The VAE was trained for 4000 epochs. Each network was trained five times and we report the average results in \cref{tab:results}. The AlexNet and the ResNet101 perform equally the best, however the AlexNet trained faster so we use it for further experiments.

\begin{table*}[ht]
    \centering
    \begin{tabular}{|l|c|c|c|c|c|}
        \hline
        & Overall ($\uparrow$) & Asphalt ($\uparrow$) & Grass ($\uparrow$) & Sand ($\uparrow$) & Sidewalk ($\uparrow$) \\
        \hline
        1D CNN \cite{Kiranyaz2021} & 0.795 & 0.921 & 0.750 & 0.721 & 0.737 \\
        Autoencoder \cite{Guo2017} & 0.873 & 0.922 & 0.967 & 0.653 & 1.000 \\
        ResNet101 \cite{He2015} & \textbf{0.985} & \textbf{1.000} & \textbf{1.000} & 0.947 & \textbf{1.000} \\
        AlexNet \cite{Alex2012} & \textbf{0.985} & \textbf{1.000} & 0.995 & \textbf{0.952} & \textbf{1.000} \\
        \hline
    \end{tabular}
    \caption{GPR classification network accuracy, averaged over five trained networks per architecture.}
    \label{tab:results}
\end{table*}

\subsection{Selection of Radargram Subsections}\label{sec:exp}

As previously mentioned, we sectioned out the portion of the radargram that represents the direct wave as this isolates the GPR's representation of the surface terrain. In this experiment, we seek to confirm our hypothesis that the direct wave is more informative for surface terrain classification than the reflected wave section of the radargram. With our frozen train and test radargram sets, we slice the training images into two sections: the direct wave and the reflected wave. Each image is of size $60 \times 32$ pixels. The total number of image slices in the train set 1033, and the total number in test is 278. Additionally, we train networks on the full (combined direct wave and reflected wave) $200 \times 32$ radargram slices. We use AlexNet classifiers for this experiment and report average accuracy across five trained networks for each approach.

The results are shown in \cref{tab:size_results}, where the full radargram performs marginally better than the direct wave. The AlexNet trained on the full image may be focusing on learning features in the region of the image that correspond to the direct wave, hence the similar performance between the two networks. The network trained on the direct wave slice significantly outperforms the AlexNet trained only on the reflected wave. These results are intuitive because the reflected wave represents the subsurface materials several feet deep, and it may also capture features such as underground pipes or water deposits. There may be some correlation between the subsurface and surface terrains that is learned by the network, but ultimately the direct wave is a much better representation. Additionally, the validation loss for the reflected wave train set converged poorly during training. Overall, this experiment is a good indicator that the direct wave is the most important part of the radargram for surface terrain classification. Still, using the full slice provides the best performance.

\begin{table}[ht]
    \centering
    \begin{tabular}{|l|c|}
        \hline
        Radargram Input Slice & Overall Accuracy ($\uparrow$) \\
        \hline
        Reflected Wave & 0.768 \\ 
        Direct Wave & 0.985 \\
        Full Slice & \textbf{0.989} \\
        \hline
    \end{tabular}
    \caption{Comparison of AlexNet networks trained on different portions of the GPR radargram. See \cref{fig:direct-wave} for an example of direct wave and reflected wave in a radargram.}
    \label{tab:size_results}
\end{table}

\begin{figure}[t]
  \centering
  \begin{subfigure}[b]{0.18\linewidth}
    \centering
    \includegraphics[height=3cm]{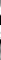}
    \caption{1 trace}
    \label{fig:sub1}
  \end{subfigure}%
  \begin{subfigure}[b]{0.18\linewidth}
    \centering
    \includegraphics[height=3cm]{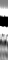}
    \caption{8 traces}
    \label{fig:sub2}
  \end{subfigure}%
  \begin{subfigure}[b]{0.18\linewidth}
    \centering
    \includegraphics[height=3cm]{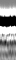}
    \caption{16 traces}
    \label{fig:sub3}
  \end{subfigure}%
  \begin{subfigure}[b]{0.18\linewidth}
    \centering
    \includegraphics[height=3cm]{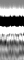}
    \caption{24 traces}
    \label{fig:sub4}
  \end{subfigure}
  \begin{subfigure}[b]{0.18\linewidth}
    \centering
    \includegraphics[height=3cm]{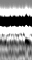}
    \caption{32 traces}
    \label{fig:sub5}
  \end{subfigure}
  \caption{The five time series lengths used for the experiment shown on a single radargram of sandy terrain. Note that the pattern distinct to the terrain type makes itself more apparent as the number of traces increases.}
  \label{fig:all}
\end{figure}

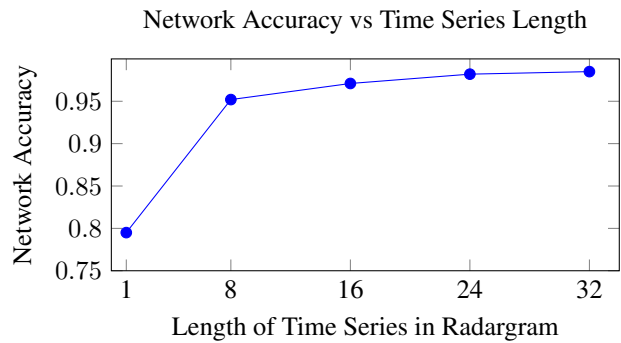
\begin{figure}[t]
    \centering
    \begin{tikzpicture}
        \begin{axis}[
            title=Network Accuracy vs Time Series Length,
            xlabel=Length of Time Series in Radargram,
            ylabel=Network Accuracy,
            xmin=0, xmax=34,
            ymin=0.75, ymax=1,
            xtick={1, 8, 16, 24, 32},
            xticklabels={1, 8, 16, 24, 32},
            ytick={0.75, 0.8, 0.85, ..., 1},
            height=4.4cm, width=\linewidth
                    ]
        \addplot[mark=*,blue] plot coordinates {
            (1, 0.795)
            (8, 0.952)
            (16, 0.971)
            (24, 0.982)
            (32, 0.985)
        };
        \end{axis}
    \end{tikzpicture}
    \caption{Averaged network accuracy plotted against the input radargram length. Notice that the performance increases minimally after time series length of 8.}
    \label{fig:timeseries}
\end{figure}

\subsection{Time Series Length}

A radargram is a time series of multiple returns from the GPR sensor stacked horizontally into an image. In this experiment, we seek to understand how many consecutive samples must be collected for the terrain to become recognizable to the classification network. In \cref{fig:all}, we see how as the width of the radargram increases, the pattern defined by the terrain becomes more apparent. However, depending on the sampling frequency, we may have to wait upwards of 15 seconds for 32 traces to be recorded -- making real-time terrain classification with this size less convenient. Additionally, passing over terrain boundaries will take longer to detect with a longer radargram. The smaller the radargram necessary for accurate classification, the more fine-grained and spatially precise our terrain predictions. We train a 1D CNN on just single traces, and also train four different AlexNets on radargrams with 8, 16, 24, and 32 stacked traces. Each network is trained five times, and the average results are presented in \cref{fig:timeseries}.

As shown in \cref{fig:timeseries}, the radargram with 32 traces performs the best, as we might expect. However, the radargrams of lengths 8, 16, and 24 perform only marginally worse, making these viable alternatives that can also provide more granular predictions. The 1D CNN, trained only on individual vector signal returns from the GPR, performs much worse -- indicating that terrain class in GPR direct wave radargrams is more detectable over multiple traces.

\begin{figure}[b]
    \centering
    \includegraphics[width=\linewidth]{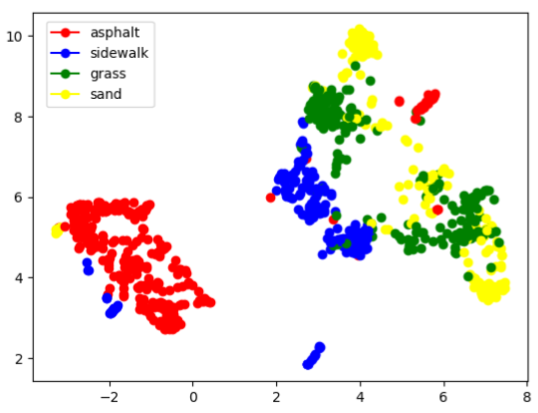}
    \caption{Deep clustering VAE results with combined loss function. Notice the distinct clusters for asphalt and sidewalk, and the more mixed clusters for sand and grass.}
    \label{fig:cluster}
\end{figure}

\subsection{Deep Clustering}

The VAE deep clustering approach to classification is unsupervised, so we may expect less accurate results compared to the supervised AlexNet and ResNet101. In \cref{fig:cluster}, we see distinct clusters for asphalt and sidewalk, and two more mixed clusters for grass and sand. The individual terrain class accuracies presented in \cref{tab:results} similarly reflect that the networks have a harder time learning the difference between sand and grass. This may be due to stronger similarities between radargrams of these two types.

\subsection{Future Applications to Mapping} \label{sec:mapping}

We conclude that GPR shows promise as an additional exteroceptive modality for terrain classification and semantic mapping. In this section, we show preliminary results that build off of the Bayesian approach in \cite{Ewen2022} to fuse GPR classification network outputs and RGB semantic segmentation network outputs into a labeled terrain map. We compare our fused vision and GPR approach against a vision-only terrain segmentation network from \cite{Zhang2018}. 

In \cref{fig:comp}, preliminary results on a test site where the robot traversed a sidewalk are shown. Of particular note is that the vision-only approach misidentifies the sidewalk (blue) as asphalt (red). In our approach, which fuses vision and GPR data, the GPR is able to successfully identify the sidewalk and correct the map semantics as the robot drives over the terrain.

%% file: sec/5_discussion.tex
\begin{figure}[t]
    \centering
    \includegraphics[width=\linewidth]{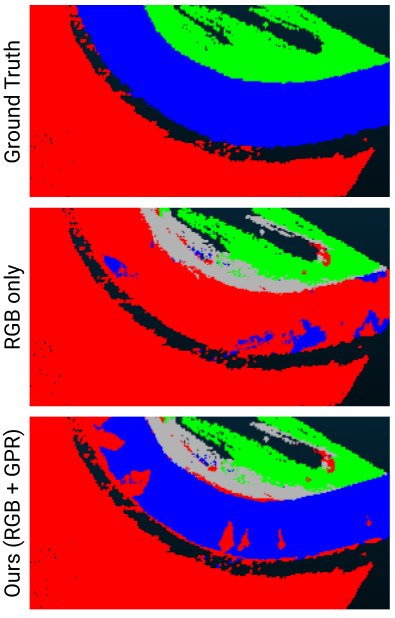}
    \caption{Bird's eye view terrain mapping comparison between ground truth, the RGB-only baseline, and our method (RGB and GPR) on a test site. Red represents asphalt, blue represents sidewalk, green represents grass, gray represents other, and black represents no data. Notice in our approach that the GPR is able to correct for the RGB camera network's misclassification of the sidewalk.}
    \label{fig:comp}
\end{figure}

\section{Discussion}
\label{sec:discussion}

Our results present a preliminary case for the use of GPR for surface terrain classification. Through the experiments presented, we show how different radargram subsections and lengths affect classification network performance. With our dataset and choice of networks, the direct wave portion of the sensor return and a time series length of at least 8 pixels provides optimal prediction results. We believe that the results presented demonstrate great promise in the novel use of GPR for surface terrain classification.

Alternative sensor modalities such as microphones, IMUs, and torque sensors are gaining increasing traction as advances in multimodal learning demonstrate their capability to compensate for the weaknesses of more traditional sensing modalities such as RGB and depth. GPR is typically a downward facing sensor, and in order to collect samples the sensor must pass directly over the terrain. As the sensor is usually not forward-looking, this limitation motivates a multimodal approach to navigation stacks that include GPR and forward-facing cameras. We show preliminary results from an RGB and GPR  fusion terrain classification and mapping framework, demonstrating that GPR can help correct for misclassifications from forward-facing camera-based networks. Additionally, it is of note that labeling GPR traces is trivial compared to other modalities, reducing the manual labeling burden. We hope to motivate applications with robotic platforms that already have a GPR to incorporate this sensor as an additional modality for the task of terrain classification.

%% file: sec/6_conclusion.tex
\section{Conclusion}
\label{sec:conclusion}

Mobile systems are cost and power constrained, making it more effective to utilize existing sensors rather than include new ones. In this work, we propose the novel use of GPR for surface terrain classification. We explore supervised and unsupervised networks for classification, and we also perform several experiments to learn more about how the radargram size and direct wave impact the classification network performance. Lastly, we demonstrate preliminary results of integrating GPR terrain classifications for a multimodal mapping framework. This motivates future work on leveraging GPR to complement RGB for terrain classification and semantic mapping.

%% file: sec/7_acknowledgements.tex
\section*{Acknowledgements}
\label{sec:acknowledgements}

This work is supported by the National Science Foundation Grant \#DGE-2241144 and a University of Michigan Space Institute Pathfinder Grant. We would like to extend special thanks to Peter Gaskell and Jason Corso for providing access to the Clearpath Husky robot. Additionally, we would like to thank Onur Bagoren, Jingyu Song, Advaith Sethuraman, Razan Andigani, Robin Wyllie-Scholz, and Hope Hanna-Casupang for assisting with data collection. We are grateful to the University of Michigan MCity for allowing us to use their facilities for data collection.